\documentclass[10pt,twocolumn,letterpaper]{article}

\usepackage{iccv}
\usepackage{times}
\usepackage{epsfig}
\usepackage{graphicx}
\usepackage{amsmath}
\usepackage{amssymb}
\usepackage{multirow}
\usepackage{booktabs}

\usepackage[pagebackref=true,breaklinks=true,letterpaper=true,colorlinks,bookmarks=false]{hyperref}

\iccvfinalcopy 

\def\iccvPaperID{****} 

\ificcvfinal\fi

\begin{document}

\title{Group-Free 3D Object Detection via Transformers}
\author{
Ze~Liu$^{1,2}$\textsuperscript{\thanks{This work is done when Ze Liu is an intern at MSRA. }}
\quad Zheng~Zhang$^2$\textsuperscript{\thanks{Contact person}} 
\quad Yue~Cao$^2$
\quad Han~Hu$^2$ 
\quad Xin~Tong$^2$ \\
{$^1$University of Science and Technology of China} \\
\small{\texttt{liuze@mail.ustc.edu.cn}} \\
{$^2$Microsoft Research Asia} \\
\small{
\texttt{\{zhez,yuecao,hanhu,xtong\}@microsoft.com}}
}

\maketitle
\ificcvfinal\thispagestyle{empty}\fi

\begin{abstract}
Recently, directly detecting 3D objects from 3D point clouds has received increasing attention. To extract object representation from an irregular point cloud, existing methods usually take a point grouping step to assign the points to an object candidate so that a PointNet-like network could be used to derive object features from the grouped points. However, the inaccurate point assignments caused by the hand-crafted grouping scheme decrease the performance of 3D object detection. 
   
In this paper, we present a simple yet effective method for directly detecting 3D objects from the 3D point cloud. Instead of grouping local points to each object candidate, our method computes the feature of an object from all the points in the point cloud with the help of an attention mechanism in the Transformers~\cite{vaswani2017attention}, where the contribution of each point is automatically learned in the network training. With an improved attention stacking scheme, our method fuses object features in different stages and generates more accurate object detection results. With few bells and whistles, the proposed method achieves state-of-the-art 3D object detection performance on two widely used benchmarks, ScanNet V2 and SUN RGB-D. The code and models are publicly available at~\url{https://github.com/zeliu98/Group-Free-3D}
\end{abstract}

\section{Introduction}
3D object detection on point cloud simultaneously localizes and recognizes 3D objects from a 3D point set. As a fundamental technique for 3D scene understanding, it plays an important role in many applications such as autonomous driving, robotics manipulation, and augmented reality. 

\begin{figure}
    \centering
    \includegraphics[width=0.9\linewidth]{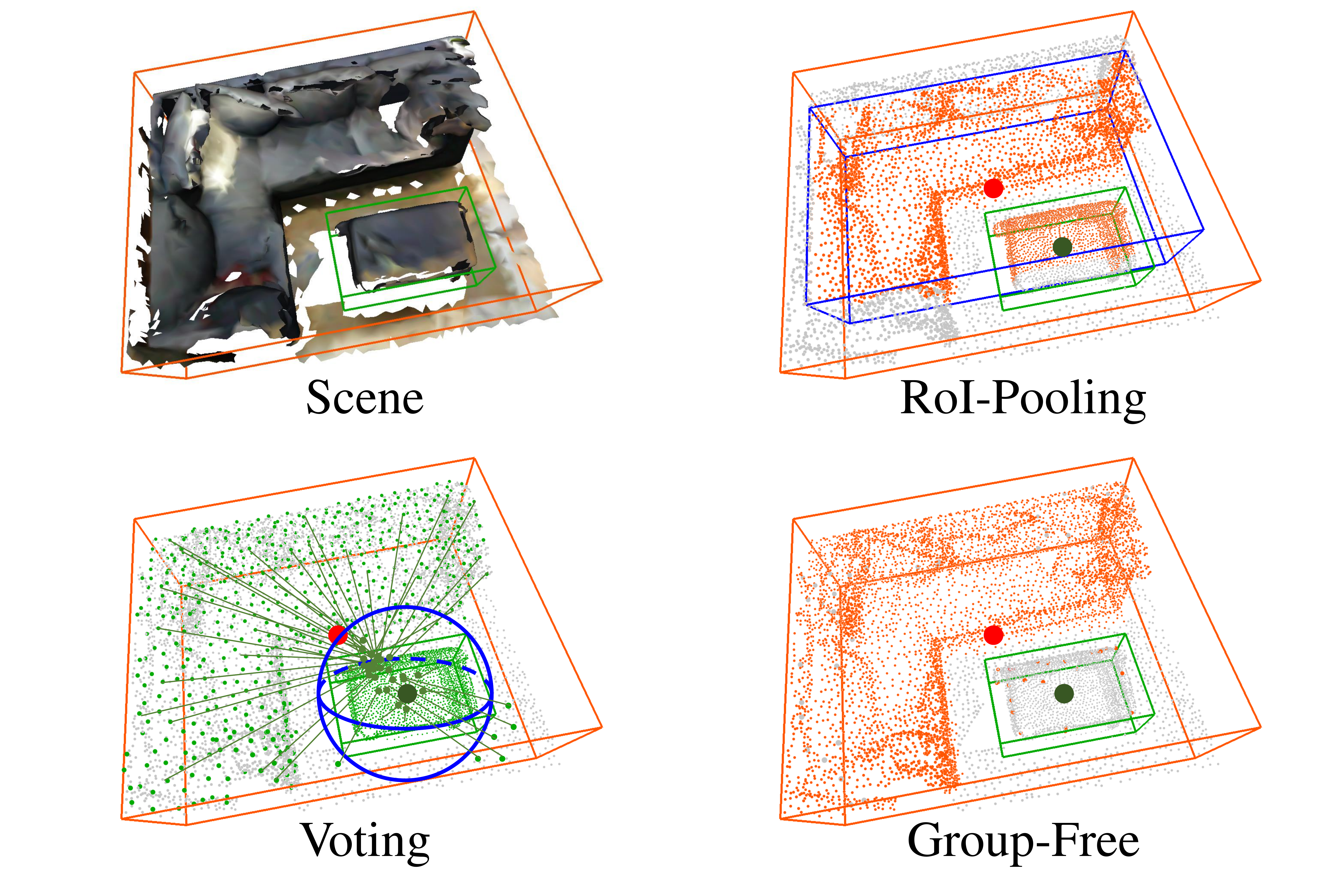}
    \caption{With the heuristic point grouping step, all points in blue box of RoI-Pooling or blue ball of Voting are assigned and aggregated to derive the object features, resulting in wrong assignments. Our group-free based approach automatically learn the contribution of all points to each object, which has ability to alleviate the drawbacks of the hand-crafted grouping.
    }
    \label{fig:teaser}
    \vspace{-1em}
\end{figure}

Different from 2D object detection that works on 2D regular images, 3D object detection takes irregular and sparse point cloud as input, which makes it difficult to directly apply techniques used for 2D object detection techniques.  
Recent studies~\cite{qi2018frustum,shi2019pointrcnn,qi2019votenet,zhang2020h3dnet} infer the object location and extract object features directly from the irregular input point cloud for object detection. In these methods, a point grouping step is required to assign a group of points to each object candidate, and then computes object features from assigned groups of points. For this purpose, different grouping strategies have been investigated. Frustum-PointNet~\cite{qi2018frustum} applies the Frustum envelop of a 2D proposal box for point grouping. Point R-CNN~\cite{shi2019pointrcnn}  groups points within the 3D box proposals to objects. VoteNet~\cite{qi2019votenet} determines the group as the points which vote to the same (or spatially-close) center point. 
Although these hand-crafted grouping schemes facilitate 3D object detection, the complexity and diversity of objects in real scene may lead to wrong point assignments (shown in Figure.~\ref{fig:teaser}) and degrade the 3D object detection performance.

In this paper, we propose a simple yet effective technique for detecting 3D objects from point clouds without the handcrafted grouping step. The key idea of our approach is to take all points in the point cloud for computing features for each object candidate, in which the contribution of each point is determined by an automatically learned attention module. Based on this idea, we adapt the Transformer to fit for 3D object detection, which could simultaneously model the object-object and object-pixel relationships, and extract the object features without handcrafted grouping.

To further release the power of the transformer architecture, we improve it in two aspects. 
First, we propose to iteratively refine the prediction of objects by updating the spatial encoding of objects in different stages, while the original application of Transformers adopt the fixed spatial encoding.
Second, we use the ensemble of detection results predicted at all stages during inference, instead of only using the results in the last stage as the final results. These two modifications efficiently improve the performance of 3D object detection with few computational overheads.

We validate our method with both ScanNet V2~\cite{dai2017scannet} and SUN RGB-D~\cite{zhou2014learning} benchmarks. Results show that our method is effective and robust to the quality of initial object candidates, where even a simple farthest point sampling approach has been able to produce strong results on ScanNet V2 and SUN RGB-D benchmarks. For the SUN RGB-D dataset, our method with the ensemble scheme results in significant performance improvement (+3.8 mAP@0.25). With few bells and whistles, the proposed approach achieved state-of-the-art performance on both benchmarks.

We believe that our method also advocates a strong potential by using the attention mechanism or Transformers for point cloud modeling, as it naturally addresses the intrinsic irregular and sparse distribution problems encountered by 3D point clouds. This is contrary to 2D image modeling, where such modeling tools mainly act as a challenger or a complementary component to the mature grid modeling tools such as ConvNets variants~\cite{hu2019local,ramachandran2019stand,wang2018non} and RoI Align~\cite{carion2020detr,chi2020relationnet++}.

\section{Related Work}

\paragraph{Grid Projection/Voxelization based Detection} Early 3D object detection approaches project point cloud to 2D grids or 3D voxels so that the advanced convolutional networks can be directly applied. A set of methods~\cite{ku2018avod,liang2018deep,yang2018pixor} project point cloud to the bird’s view and then employ 2D ConvNets for learning features and generate 3D boxes. These methods are mainly applied for the outdoor scenes in autonomous driving where objects are distributed on a horizontal plane so that their projections on the bird-view are occlusion-free. Note these approaches also need to address the irregular and sparse distribution issues of the 2D point projections, usually by pixelization. Other methods~\cite{chen2017multi,xu2018multi} project point clouds into frontal views and then apply 2D ConvNets for object detection. Voxel-based methods~\cite{song2016deep,zhou2018voxelnet} convert points into voxels and employ 3D ConvNets to generate features for 3D box generation. All these projection/voxelization based methods suffer from quantization errors. The voxel-based methods also suffer from the large memory and computational cost of 3D convolutions.

\paragraph{Point based Detection} Recent methods directly process point clouds for 3D object detection. A core task of these methods is to compute object features from the irregularly and sparsely distributed points. All existing methods first assign a group of points to each object candidate and then compute object features from each point group. Frustum-PointNet~\cite{qi2018frustum} groups points by the 3D Frustum envelope of a 2D box detected using an RGB object detector, and applies a PointNet on the grouped points to extract object features for 3D box prediction. Point R-CNN~\cite{shi2019pointrcnn} directly computes 3D box proposals, where the points within this 3D box are used for object feature extraction. PV-RCNN~\cite{shi2020pv} leverages the voxel representation to complement the point-based representation in Point R-CNN~\cite{shi2019pointrcnn} for 3D object detection and achieves better performance.

VoteNet \cite{qi2019votenet} groups points according to their voted centers and extract object features from grouped points by the PointNet. Some follow-up works further improve the point group generation procedure \cite{zhang2020h3dnet} or the object box localization and recognition procedure \cite{chen2020hierarchical}. 

Our method is also a point-based detection approach. Unlike existing point-based approaches, our method involves all the points for computing the features of each object candidate by an attention module. We also stack the attention modules to iteratively refine the detection results while maintaining the simplicity of our method. 

\paragraph{Network architecture for Point Cloud} A large set of network architectures ~\cite{su2015multi,guo2016multiview3d,qi2016volumetric,feng2018gvcnn,wu20153d,maturana2015voxnet,wang2017ocnn,tatarchenko2017octree,wang2018adaptive, qi2017pointnet,qi2017pointnet++,shen2018mining,wang2018local,te2018rgcnn,jampani2016learning,atzmon2018point,thomas2019kpconv,xu2018spidercnn,groh2018flex,liu2020closer} have been proposed for various point cloud based learning tasks. \cite{guo2020deep} provides a good taxonomy and review of all these architectures, and discussing all of them is beyond the scope of this paper. Our method can take any point cloud architecture as the backbone network for computing the point features. We adopt PointNet++~\cite{qi2017pointnet++} used in previous methods~\cite{qi2019votenet,qi2020imvotenet,zhang2020h3dnet} in our implementation for a fair comparison.

\begin{figure*}
    \centering
    \includegraphics[width=0.85\linewidth]{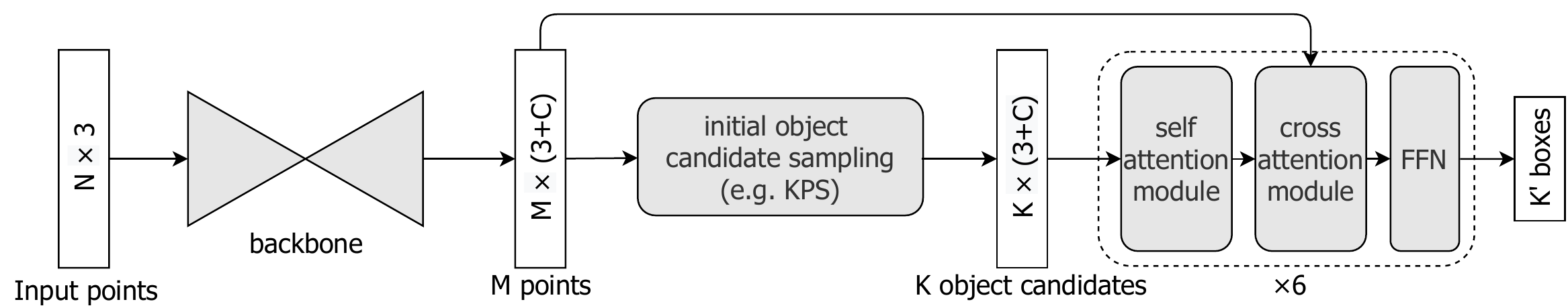}
    \caption{This figure illustrates the simple architecture of our approach, including three major components: a {backbone network} to extract feature representations for each point in the point cloud, a {sampling method} to generate initial object candidates, and {stacked attention modules} to extract and refine object representations from all points.}
    \label{fig:framework}
    \vspace{-1em}
\end{figure*}

\paragraph{Attention Mechanism/Transformer in NLP and 2D Image Recognition} 
The attention-based Transformer is the dominant network architecture for the learning tasks in the field of NLP~\cite{vaswani2017attention,devlin2018bert,liu2019roberta}. They have been also applied in the field of 2D image recognition~\cite{hu2019local,ramachandran2019stand,wang2018non} as a strong competitor to the dominant grid/dense modeling tools such as ConvNets and RoI-Align. The most related works in 2D image recognition to this paper are those who apply the attention mechanism or Transformer architectures into 2D object detection~\cite{hu2018relation,gu2018learning,chi2020relationnet++,carion2020detr}. 

Among these approaches, our method is most similar to \cite{carion2020detr}, which also applies a Transformer architecture for 2D object detection. However, we found that directly applying this method to point clouds leads to significantly lower performance than our approach in 3D object detection task. On the one hand, this is caused by the new technologies we proposed, and on the other hand, it probably because our method better integrated the advantage of traditional 3D detection framework. We discussed these factors in Sec.~\ref{sec:DETR}.

Our approach improves the Transformer models to better adapt the 3D object detection task, including the update of object query locations in the multi-stage iterative box prediction, and an ensemble of detection results of stages.
Although the attention mechanisms still have a certain performance gap compared to the dominant convolution-based methods in other tasks, we found that this architecture may well address the point grouping issue for object detection on point clouds. As a result, we advocate a strong potential of this architecture for modeling irregular 3D point clouds.

\section{Methodology}

In 3D object detection on point clouds, we are given a set of $N$ points $S \in \mathbb{R}^{N\times 3}$ and the goal is to produce a set of 3D (oriented) bounding boxes with categorization scores $\mathcal{O}_S$ to cover all ground-truth objects. Our overall architecture is illustrated in Figure~\ref{fig:framework}, involving three major components: a \emph{backbone network} to extract feature representations for each point in point clouds, a \emph{sampling method} to generate initial object candidates, and \emph{stacked attention modules} to extract and refine object representations from all points.

\paragraph{Backbone Architecture} While our framework can leverage any point cloud network to extract point features, we adopt PointNet++~\cite{qi2017pointnet++} as the backbone network for a fair comparison with the recent methods~\cite{qi2019votenet,zhang2020h3dnet}.

The backbone network receives a point cloud of $N$ points (i.e. 2048) as input. We follow the encoder-decoder architecture in ~\cite{qi2017pointnet++} to first down-sample the point cloud input into $8\times$ resolution (i.e. 256 points) through four stages of set abstraction layers, and then up-sample it to the resolution of $2\times$ (i.e. 1024 points) by feature propagation layers. The network will produce a $C$-channel vector representation for each point on the $2\times$ resolution, denoted as $\{{\bf z}_i\}_{i=1}^{M}$, which are then used in the \emph{initial object candidates sampling} module and the \emph{stacked attention} modules. In the following parts, we will first describe these two modules in detail, and then present the loss function and head design for this framework.

\subsection{Initial Object Candidate Sampling}
\label{sec:sampling}

While object detection on 2D images usually adopts data-independent anchor boxes as initial object candidates, it is generally intractable or impractical for 3D object detection to apply this simple top-down strategy, as the number of anchor boxes in 3D search space is too huge to handle. Instead, we follow recent practice \cite{shi2019pointrcnn,qi2019votenet} to sample initial object candidates directly from the points on a point cloud, by a bottom-up way. 

We consider three simple strategies to sample initial object candidates from a point cloud:
\begin{itemize}
    \item \emph{Farthest Point Sampling (FPS)}. The FPS approach has been widely adopted to generate a point cloud from a 3D shape or to down-sample the point clouds to a lower resolution. This method can be also employed to sample initial candidates from a point cloud. Firstly, a point is randomly sampled from the point cloud. Then the farthest point to the already-chosen point set is iteratively selected until the number of chosen points meets the candidate budget. Though it is simple, we show in experiments that this sampling approach along with our framework has been able to be comparable to the previous state-of-the-art 3D object detectors.
    \item \emph{$k$-Closest Points Sampling (KPS)}. In this approach, we classify each point on a point cloud to be a real object candidate or not. The label assignment in training follows this rule: a point is assigned positive if it is inside a ground-truth object box and it is one of the $k$-closest points to the object center. In inference, the initial candidates are selected according to the classification score of the point.

    \item \emph{KPS with non-maximal suppression (KPS-NMS)}. Built on the above \emph{KPS} method, we introduce an additional non-maximal suppression (NMS) step, which iteratively removes spatially close object candidates, to improve the recall of sampled object candidates given a fixed number of objects, following the common practice in 2D object detection. In addition to the \emph{objectness} scores, we predict also the object center that each point belongs to, where the NMS is conducted accordingly. Specifically, the candidates locating within a radius of the selected object center will be suppressed. The radius is set to 0.05 in our experiments.
\end{itemize}

In experiments, we will demonstrate that our framework has strong compatibility with the choice of these sampling approaches, mainly ascribed to the robust object feature extraction approach described in the next subsection (see Table~\ref{tab::ablation_sampling}). \textbf{We use the \emph{KPS} approach by default}, due to its better performance than the \emph{FPS} approach, and the same effectiveness as the more complex \emph{KPS-NMS} approach.

\subsection{Iterative Object Feature Extraction and Box Prediction by Transformer Decoder}

With the initial object candidates generated by a sampling approach, we adopt the Transformer as the decoder to leverage all points on a point cloud to compute the object feature of each candidate. The multi-head attention network is the foundation of Transformer, it has three input sets: query set, key set and value set. Usually, the key set and value set are different projections of the same set of elements. Given a query set $\{\mathbf{q}_i\}$ and a common element set $\{\mathbf{p}_k\}$ of key set and value set, the output feature of the multi-head attention of each query element is the aggregation of the values that weighted by the attention weights, formulated as:
\begin{equation}
\label{eq:att}
\text{Att}(\mathbf{q}_{i}, \{\mathbf{p}_{k}\}) = \sum_{h=1}^{H} W_h  ( \sum_{k=1}^{K} A^{h}_{i,k} \cdot V_h {\mathbf{p}}_k),
\end{equation}

\begin{equation}
\label{eq:att_weight}
A^{h}_{i,k} = \frac{\exp[(Q_h \mathbf{q}_{i})^T (U_h \mathbf{p}_k)]}{\sum_{k=1}^{K}{\exp[(Q_h \mathbf{q}_{i})^T (U_h \mathbf{p}_k)]}}
\end{equation}
where $h$ indexes over attention heads, $A_h$ is the attention weight, 
$Q_h, V_h, U_h, W_h$ indicate the query projection weight, value projection weight, key projection weight, and output projection weight, respectively. 

\begin{figure}
    \centering
    \includegraphics[width=0.5\columnwidth]{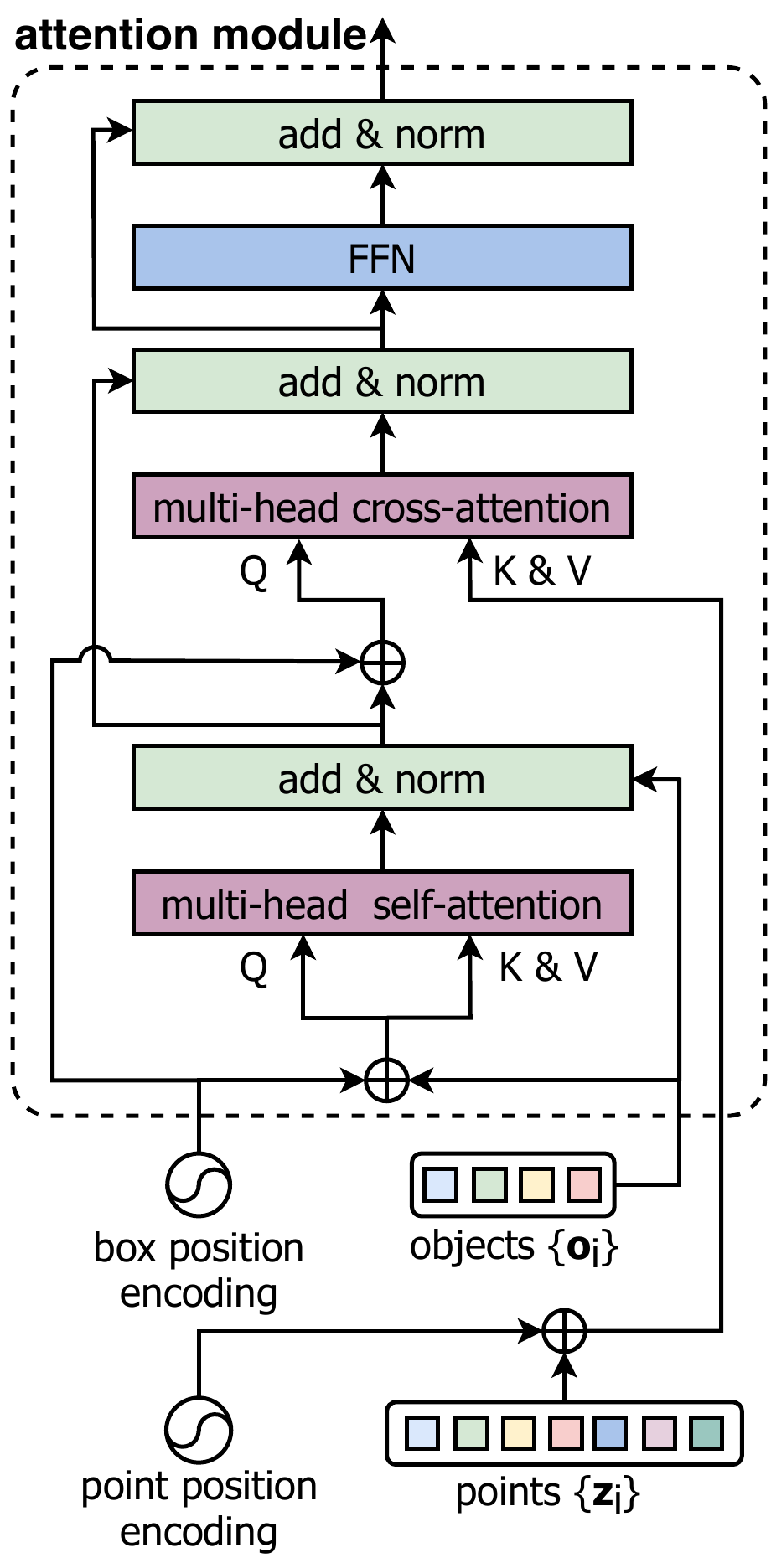}
    \caption{Architecture of the attention module.}
    \label{fig:att_mod}
    \vspace{-1em}
\end{figure}

While the standard Transformer predicts the sentence of a target language sequentially in an auto-regressive way, our Transformer computes object features and predicts 3D object boxes in parallel. The Transformer consists of several stacked multi-head \textit{self-attention} and multi-head \textit{cross-attention} modules, as illustrated in Figure~\ref{fig:att_mod}. 

Denote the input point features at stage $l$ as $\{{\mathbf{z}}_i^{(l)}\}_{i=1}^{M}$ and the object features at the same stage as $\{{\mathbf{o}}_i^{(l)}\}_{i=1}^{K}$. A self-attention module models interaction between object features, formulated as:
\begin{equation}
\label{eq:self_att}
\text{Self-Att}({\mathbf{o}}_i^{(l)},\{\mathbf{o}_j^{(l)}\}) = \text{Att}({\mathbf{o}}_i^{(l)},\{\mathbf{o}_j^{(l)}\}),
\end{equation}
A cross-attention module leverages point features to compute object features, formulated as:

\begin{equation}
\label{eq:cross_att}
\text{Cross-Att}({\mathbf{o}}_i^{(l)},\{\mathbf{z}_j^{(l)}\})=\text{Att}({\mathbf{o}}_i^{(l)},\{\mathbf{z}_j^{(l)}\}),
\end{equation}
where the notations are similar to those in Eq.~(\ref{eq:self_att}). After the object feature are updated through the self-attention module and cross attention module, a feed-forward network (FFN) is then applied to further transformed feature of each object. 

There are a few differences compared to the original Transformer decoders, as described below.

\paragraph{Iterative Object Box Prediction and Spatial Encoding} The original Transformer adopts a fixed spatial encoding for all of the stacked attention modules, indicating the indices of each word. The application of Transformers to 2D object detection \cite{carion2020detr} instantiate the spatial encoding (object prior) as a learnable weight. During inference, the spatial encoding is fixed and same for any images.

In this work, we propose to refine the spatial encodings of an object candidate stage by stage. Specifically, we predict the 3D box locations and categories at each decoder stage, and the predicted location of a box in one stage will be used to produce the refined spatial encoding of the same object, the refined spatial encoding vector is then added to the output feature of this decoder stage and fed into the next stage. The spatial encodings of an object and a point are computed by applying independent linear layers on the parameterization vector of a 3D box $(x,y,z,l,h,w)$ and a point $(x,y,z)$, respectively. In the experiments, we will show this approach can improve the mAP@0.25 and mAP@0.5 by 1.6 and 5.0 on the ScanNet V2 benchmark, compared to the approach without iterative refinement. 

\paragraph{Ensemble from Multi-Stage Predictions} 
Another difference is that we ensemble the predictions of different stages to produce final detection results, while previous methods usually adopt the output of the last stage as the final results. Concretely, the detection results of different stages are combined and they together go through an NMS (IoU threshold of 0.25) procedure to generate the final object detection results. We find this approach can significantly improve the performance of some benchmarks, e.g. +3.8 mAP@0.25 on the SUN RGB-D dataset. Also note the overhead of this ensembling approach is marginal, mainly ascribed to the multi-stage nature of the Transformer decoder.

\subsection{Heads and Loss Functions}
\paragraph{Decoder Head}
We apply head networks on all decoder stages, with each mostly following the setting in~\cite{qi2019votenet}. There are 5 prediction tasks: objectness prediction with a binary focal loss~\cite{lin2017focal} $\mathcal{L}_\text{obj}$, box classification with a cross entropy loss $\mathcal{L}_\text{cls}$, center offset prediction with a smooth-L1 loss $\mathcal{L}_\text{center\_off}$, size classification with a cross entropy loss $\mathcal{L}_\text{sz\_cls}$, and size offset prediction with a smooth-L1 loss $\mathcal{L}_\text{sz\_off}$. 
Also, all 5 prediction tasks are obtained by a shared 2-layer MLP and an independent linear layer. 

The loss of $l$-th decoder stage is the combination of these 5 loss terms by weighted summation:
\begin{equation}
    \mathcal{L}_\text{decoder}^{(l)} = \beta_1\mathcal{L}_\text{obj}^{(l)} + \beta_2\mathcal{L}_\text{cls}^{(l)} + \beta_3  \mathcal{L}_\text{center\_off}^{(l)} + \beta_4 \mathcal{L}_\text{sz\_cls}^{(l)} + \beta_5 \mathcal{L}_\text{sz\_off}^{(l)},
\end{equation}
where the balancing factors are set default as $\beta_1=0.5$, $\beta_2=0.1$, $\beta_3=1.0$, $\beta_4=0.1$ and $\beta_5=0.1$. The losses on all decoder stages are averaged to form the final loss:
\begin{equation}
    \mathcal{L}_\text{decoder} = \frac{1}{L}\sum _{l=1}^{L} \mathcal{L}_\text{decoder}^{(l)}.
\end{equation}

\paragraph{Sampling Head}
The head designs and the loss functions of the sampling module are similar to those of the decoders. There are two differences: firstly, the box classification task is not involved; secondly, the objectness task follows the label assignment as described in Sec.~\ref{sec:sampling}. Our final loss is the sum of decoder and sampling heads:
\begin{equation}
    \mathcal{L} = \mathcal{L}_\text{decoder} + \mathcal{L}_\text{sampler}
\end{equation}

\section{Experiments}\label{sec:exp}
\subsection{Datasets and Evaluation Protocol}
We validate our approach on two widely-used 3D object detection datasets: ScanNet V2~\cite{dai2017scannet} and SUN RGB-D~\cite{song2015sun}, and we follow the standard data splits~\cite{qi2019votenet} for them both.

\noindent \textbf{ScanNet V2~\cite{dai2017scannet}} is constructed from an 3D reconstruction dataset of indoor scenes by enriched annotations. It consists of 1513 indoor scenes and 18 object categories. The annotations of per-point instance, semantic labels, and 3D bounding boxes are provided. We follow a standard evaluation protocol~\cite{qi2019votenet} by using mean Average Precision(mAP) under different IoU thresholds, without considering the orientation of bounding boxes. 

\noindent \textbf{SUN RGB-D~\cite{song2015sun}} is a single-view RGB-D dataset for 3D scene understanding, consisting of $\sim$5K indoor RGB and depth images. The annotation consists of per-point semantic labels and oriented bounding object bounding boxes of 37 object categories. The standard mean Average Precision is used as evaluation metrics and the evaluation is reported on the 10 most common categories, following~\cite{qi2019votenet}.

\begin{table*}
\small
\begin{center}
\begin{tabular}{c|c|c|c}
\hline
methods & backbone & mAP@0.25 & mAP@0.5 \\
\hline
\hline
HGNet~\cite{chen2020hierarchical} & GU-net & 61.3 & 34.4 \\
GSDN~\cite{gwak2020generative} & MinkNet & 62.8 & 34.8 \\
3D-MPA~\cite{engelmann20203d} & MinkNet & 64.2 & 49.2 \\
VoteNet~\cite{qi2019votenet}\footnotemark[2] & PointNet++ & 62.9 & 39.9 \\
MLCVNet~\cite{xie2020mlcvnet} & PointNet++ & 64.5 & 41.4 \\
H3DNet~\cite{zhang2020h3dnet} & PointNet++ & 64.4 & 43.4 \\
H3DNet~\cite{zhang2020h3dnet} & 4$\times$PointNet++ & 67.2 & 48.1 \\
\hline
Ours (L6, O256) & PointNet++ & 67.3 (66.3)& 48.9 (48.5) \\
Ours (L12, O256) & PointNet++ & 67.2 (66.6) & 49.7 (49.0)\\
Ours (L12, O256) & PointNet++w2$\times$ & 68.8 (67.7) & 52.1 (50.6)\\
Ours (L12, O512) & PointNet++w2$\times$ & \textbf{69.1} (68.6) & \textbf{52.8} (51.8)\\
\hline
\end{tabular}
\end{center}
\vspace{-0.5em}
\caption{System level comparison on ScanNet V2 with state-of-the-arts. The main comparison is based on the best results of multiple experiments between different methods, and the number within the bracket is the average result.\\
\footnotesize{\textbf{Notations:} 4$\times$PointNet++ denotes 4 individual PointNet++; PointNet++w2$\times$ denotes the backbone width is expanded by 2 times; L denotes the decoder depth, and O denotes the number of object candidates, e.g. Ours (L6, O256) denotes a model with 6-layer decoder(i.e. 6 attention modules) and 256 object candidates.}\\
}
\label{tab::system_scannet_v2}
\vspace{-1em}
\end{table*}

\begin{table*}
\small
\begin{center}
\begin{tabular}{c|c|c|c|c}
\hline
methods & backbone & inputs & mAP@0.25 & mAP@0.5 \\
\hline
\hline
VoteNet~\cite{qi2019votenet}\footnotemark[2] & PointNet++ & point & 59.1 & 35.8\\
MLCVNet~\cite{xie2020mlcvnet} & PointNet++ & point & 59.8 & - \\
HGNet~\cite{chen2020hierarchical} & GU-net & point & 61.6 & - \\
H3DNet~\cite{zhang2020h3dnet} & 4$\times$PointNet++ & point & 60.1 & 39.0 \\
imVoteNet~\cite{qi2020imvotenet}$^{*}$ & PointNet++ & point+RGB & \textbf{63.4} & -\\
\hline
Ours (L6, O256) & PointNet++ & point & 63.0 (62.6) & \textbf{45.2} (44.4) \\
\hline
\end{tabular}
\end{center}
\vspace{-0.5em}
\caption{System level comparison on SUN RGB-D with state-of-the-arts. The main comparison is based on the best results of multiple experiments between different methods, and the number within the bracket is the average result. $^{*}$imVoteNet use RGB images as addition inputs.}
\label{tab::system_sunrgbd}
\vspace{-1em}
\end{table*}

\subsection{Implementation Details}
 \vspace{5pt}
\noindent \textbf{ScanNet V2}\hspace{2pt} We follow recent practice~\cite{qi2019votenet,xie2020mlcvnet} to use PointNet++ as default backbone network for a fair comparison. The backbone has 4 set abstraction layers and 2 feature propagation layers. For each set abstraction layer, the input point cloud is sub-sampled to 2048, 1024, 512, and 256 points with the increasing receptive radius of 0.2, 0.4, 0.8, and 1.2, respectively. Then, two feature propagation layers successively up-sample the points to 512 and 1024. More training details are given in Appendix.

\noindent \textbf{SUN RGB-D} The implementation mostly follow~\cite{qi2019votenet}. We use 20k points as input for each point cloud. The network architecture and the data augmentation are the same as that for ScanNet V2. As the orientation of the 3D box is required in evaluation, we include an additional orientation prediction branch for all decoder stages. More training details are given in Appendix.

\subsection{System-level Comparison}
In this section, we compare with previous state-of-the-arts on ScanNet V2 and SUN RGB-D. Since previous works~\cite{qi2019votenet, MMDetection3D} usually report the best results of multiple times on training and testing in the system-level comparison, we report both best results and average results\footnote[1]{We train each setting 5 times and test each training trial 5 times. The average performance of these 25 trials is reported to account for algorithm randomness.}

\vspace{5pt}
\noindent \textbf{ScanNet V2}\hspace{2pt} The results are shown in Table~\ref{tab::system_scannet_v2}. With the same backbone network of a standard PointNet++, the proposed approach achieves 67.3 mAP@0.25 and 48.9 mAP@0.5 using 6 decoder stages and 256 object candidates, which is 2.8 and 5.5 better than previous best results using the same backbones. By more decoder stages as 12, 
the gap increases to 6.3 on mAP@0.5.

With stronger backbones and more sampled object candidates, i.e. $2\times$ more channels and 512 candidates, the performance of the proposed approach is improved to 69.1 mAP@0.25 and 52.8 mAP@0.5, outperforming previous best method by a large margin.

\vspace{5pt}
\noindent \textbf{SUN RGB-D} \hspace{2pt}
We also compare the proposed approach with previous state-of-the-arts on the SUN RGB-D dataset, which is another widely used 3D object detection benchmark. In this dataset, the ensemble approach over multiple stages is used by default during inference. The results are shown in Table.~\ref{tab::system_sunrgbd}. Our base model achieves 63.0 on mAP@0.25 and 45.2 on mAP@0.5, which outperforms all previous state-of-the-arts that only use the point cloud. In particular, it outperforms the H3DNet on mAP@0.5 by 6.2.

\footnotetext[2]{We report the results of MMDetection3D(\href{url}{https://github.com/open-mmlab/mmdetection3d}) instead of the official paper, which reported 46.8 mAP@0.25 and 24.7 mAP@0.5 on ScanNet V2, and 57.7 mAP@0.25 and 32.0 mAP@0.5 on SUN RGB-D.}

\begin{table}
\small
\begin{center}
\begin{tabular}{c|c|c}
\hline
sampling method & mAP@0.25 & mAP@0.5 \\
\hline
\hline
FPS & 64.5 & 46.2 \\
KPS-NMS & 65.8 & \textbf{48.7}\\
KPS & \textbf{66.3} & 48.5\\
\hline
\end{tabular}
\vspace{-0.5em}
\end{center}
\caption{Ablation study on applying different sampling strategies.}
\label{tab::ablation_sampling}
\vspace{-1em}
\end{table}

\subsection{Ablation Study}
In this section, we validate our key designs on ScanNet V2. If not specified, all models have 6 attention modules, 256 sampled candidates, and are equipped with the proposed iterative object prediction approach. In evaluation, we report the average performance of 25 trials by default.

\begin{table}
\small
\begin{center}
\begin{tabular}{c|c|c}
\hline
$k$ & mAP@0.25 & mAP@0.5 \\
\hline
\hline
1 & 65.7 & \textbf{48.7} \\
2 & 65.8 & 48.3 \\
4 & \textbf{66.3} & 48.5 \\
6 & 66.1 & 48.4 \\
\hline
\end{tabular}
\vspace{-0.5em}
\end{center}
\caption{Ablation study on different values of $k$ in KPS sampling strategy.}
\label{tab::ablation_topk}
\vspace{-1em}
\end{table}

 \vspace{5pt}
\noindent \textbf{Sampling Strategy} \hspace{2pt}
We first ablate the effects of different sampling strategies in Table.~\ref{tab::ablation_sampling}. It shows that our approach performs well by using different sampling strategies. It also works well in a wide range of hyper-parameters, such as $k$ in the \emph{KPS} sampling approach (see Table.~\ref{tab::ablation_topk}).

These results indicate the robustness of our framework for choosing different sampling approaches.

\begin{table}
\small
\begin{center}
\begin{tabular}{c|c|c|c}
\hline
iterative & position encoding & mAP@0.25 & mAP@0.5 \\
\hline
\hline
 & none & 64.7 & 43.4\\
 & center+size & 64.6 & 43.5\\
\hline
\checkmark & center & 65.2 & 47.5 \\
\checkmark & center+size & \textbf{66.3} & \textbf{48.5} \\
\hline
\end{tabular}
\end{center}
\vspace{-0.5em}
\caption{Ablation study on the effectiveness of iterative box prediction.}
\label{tab::ablation_iterative_performance}
\vspace{-1em}
\end{table}

 \vspace{5pt}
\noindent \textbf{Iterative Box Prediction} \hspace{2pt}
Table~\ref{tab::ablation_iterative_performance} ablates several design choices for iterative box prediction. With a naive iterative method where no spatial encoding is involved in the decoder stages, the approach shows reasonably good performance of 64.7 mAP@0.25 and 43.4 mAP@0.25, likely because the location information may have been implicitly included in the input object features. Actually, an additional fixed position encoding does not improve detection performance (64.6 mAP@0.25 and 43.5 mAP@0.5). 

By refining the encodings of the box location stage by stage, the localization ability of the approach is significantly improved of the 4.1 points gains on the mAP@0.5 metric over the naive implementation (47.5 vs. 43.4). Also, more detailed spatial encoding by both box center and size is beneficial, compared to that only encodes box centers (66.3 vs. 65.2 on mAP@0.25 and 48.5 vs. 47.5 on mAP@0.5).

\begin{table}
\small
\begin{center}
\begin{tabular}{c|c|c}
\hline
\# of layers & mAP@0.25 & mAP@0.5 \\
\hline
\hline
0 & 63.3 & 40.7 \\
1 & 64.8 & 43.9 \\
2 & 66.0 & 45.6 \\
3 & 66.4 & 46.6 \\
4 & 66.2 & 47.9 \\
5 & 66.3 & 48.3 \\
6 &  \textbf{66.3} & \textbf{48.5} \\
\hline
\end{tabular}
\vspace{-0.5em}
\end{center}
\caption{Ablation study on the performance of iterative box prediction with different decoder layers.}
\label{tab::ablation_iterative_diff_layers}
\vspace{-1em}
\end{table}

Table.~\ref{tab::ablation_iterative_diff_layers} shows the performance of iterative box prediction with different decoder stages. More stages can bring significant performance improvement, especially in the mAP@0.5. Compared with not applying any attention modules, our 6-stage model performs better on mAP@0.25 and mAP@0.5 by 3.0 and 7.8, respectively.

\begin{table}
\addtolength{\tabcolsep}{-2.2pt}
\small
\begin{center}
\begin{tabular}{c|cc|cc}
\hline
\multirow{2}{*}{ensemble} & \multicolumn{2}{c|}{ScanNet V2} & \multicolumn{2}{c}{SUN RGB-D} \\
\cline{2-5}
& mAP@0.25 & mAP@0.5 & mAP@0.25 & mAP@0.5\\
\hline
\hline
& 66.3 & 48.5 & 59.2 & 43.3\\
\checkmark & \textbf{66.4} & \textbf{48.7} & \textbf{63.0} & \textbf{45.2} \\
\hline
\end{tabular}
\end{center}
\caption{Ablation study on the effectiveness of multi-stage ensemble.}
\label{tab::ablation_ensemble}
\vspace{-1em}
\end{table}

 \vspace{5pt}
\noindent \textbf{Ensemble Multi-stage Predictions}\hspace{2pt} Each decoder stage of our approach will predict a set of 3D boxes. It is natural to ensemble these results of different decoder stages in expecting better final detection results. Table \ref{tab::ablation_ensemble} shows the results, where significantly performance improvements are observed on SUN RGB-D (+3.8 mAP@0.25 and +1.9 mAP@0.5) and maintained performance on ScanNet V2. We hypothesize that it is because the point clouds of SUN RGB-D have lower quality than those of ScanNet V2: SUN RGB-D adopts real RGB-D signals to generate point clouds that many objects have missing parts due to occlusion, while the ScanNet V2 generate point clouds from 3D shape meshes which are more complete. The ensemble method can boost the performance more on real 3D scenes. 

\begin{figure*}
    \centering
    \includegraphics[width=0.8\linewidth]{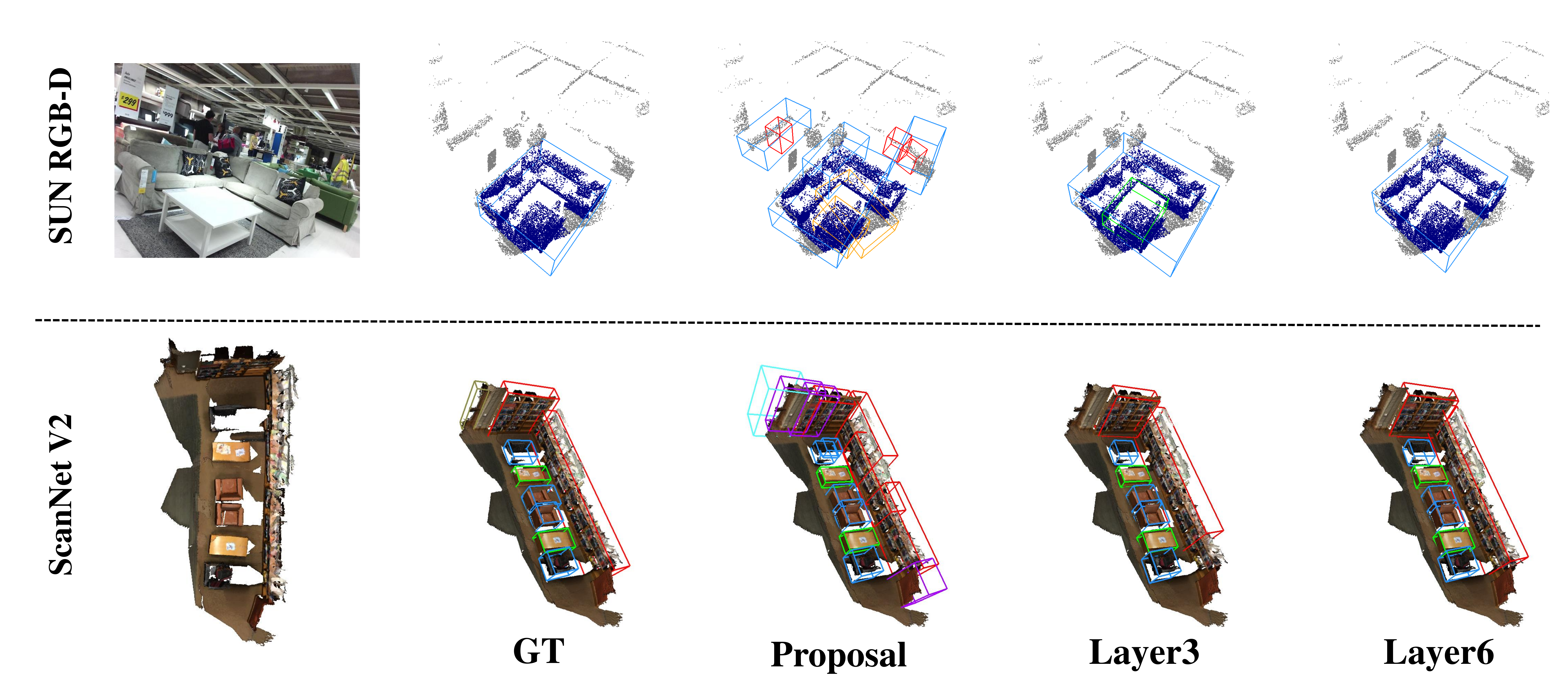}
    \caption{Qualitative results of different decoder stages. The first row is the results on SUN RGB-D, and the second row is the results on ScanNet V2. The color of bounding boxes represents their category.}
    \label{fig:qualitative}
    \vspace{-1em}
\end{figure*}

\noindent \textbf{Comparison with Group-based Approaches}
Aggregating point features through RoI-Pooing, or according to the voted centers are two typical handcrafted grouping strategies~\cite{shi2019pointrcnn, qi2019votenet} in 3D object detection. 
We refer these two grouping strategies as baselines and compare with them. For a fair comparison, we only switch the feature aggregation mechanism while all other settings (e.g. the 6-stage decoder) remain unchanged. More details are in Appendix. Table~\ref{tab::ablation_group_and_groupfree} show the results. Although RoI-Pooling outperforms than the voting approach, it is still worse than our group-free approach by 1.2 points on mAP@0.25 and 4.1 points on mAP@0.5. 
\begin{table}
\small
\begin{center}
\begin{tabular}{c|c|c}
\hline
method & mAP@0.25 & mAP@0.5 \\
\hline
\hline
RoI-Pooing & 65.1 & 44.4 \\
Voting & 64.2 & 44.1 \\
Ours & \textbf{66.3} & \textbf{48.5} \\
\hline
\end{tabular}
\end{center}
\vspace{-1em}
\caption{Comparison with grouping-based approaches.}
\label{tab::ablation_group_and_groupfree}
\vspace{-1em}
\end{table}

\begin{table}
\addtolength{\tabcolsep}{-1pt}
\small
\begin{center}
\begin{tabular}{c|c|c|c|c}
\hline
\multirow{2}{*}{method} & \multirow{2}{*}{backbone} & \multicolumn{2}{c|}{mAP} & \multirow{2}{*}{frames/s} \\
\cline{3-4}
& & 0.25 & 0.5 & \\
\hline
\hline
MLCVNet~\cite{xie2020mlcvnet} & PointNet++& 64.5 & 41.4 & 5.44 \\
H3DNet~\cite{zhang2020h3dnet} & 4$\times$PointNet++& 67.2 & 48.1 & 3.76 \\
\hline
Ours (L6, O256) & PointNet++& 67.3 & 48.9 & \textbf{6.71} \\
Ours (L12, O256) & PointNet++ & 67.2 & 49.7 & 5.70\\
Ours (L12, O256) & PointNet++w2$\times$& 68.8 & 52.1 & 5.23\\
Ours (L12, O512) & PointNet++w2$\times$& \textbf{69.1} & \textbf{52.8} & 5.17\\
\hline
\end{tabular}
\end{center}
\vspace{-0.5em}
\caption{Comparison on realistic inference speed on ScanNet V2.}
\label{tab::system_inference_speed}
\vspace{-1em}
\end{table}

\subsection{Inference Speed}
The computational complexity of the attention model is determined by the number of points in a point cloud and the number of sampled object candidates. In our approach, only a small number of object candidates are sampled, which makes the cost of the attention model insignificant. With our default setting (256 object candidates, 1024 output points), stacking one attention model brings 0.95 GFLOPs, which is quite light compared to the backbone.

In addition, the realistic inference speed of our method is also very competitive, compared to other state-of-the-art methods. For a fair comparison, all experiments are run on the same workstation (single Titan-XP GPU, 256G RAM, and Xeon E5-2650 v3) and environment (Ubuntu-16.04, Python 3.6, Cuda-10.1, and PyTorch-1.3.1). The official code of other methods is used for evaluation. The batch size of all experiments is set to 1 (i.e. single image). The results are shown in Table.~\ref{tab::system_inference_speed}. Our method achieves better performance and also higher inference speed. 

\begin{table}
\small
\begin{center}
\begin{tabular}{l|c|c|c}
\hline
method & epoch & mAP@0.25 & mAP@0.5 \\
\hline
\hline
DETR & 400 & 39.6 & 21.4\\
DETR+KPS & 400 & 59.6 & 41.0 \\
DETR+KPS+iter pred & 400 & 59.9 & 42.9 \\
DETR+KPS+iter pred & 1200 & 61.8 & 45.2 \\
\hline
Ours & 400 & \textbf{66.3} & \textbf{48.5} \\
\hline
\end{tabular}
\vspace{-0.5em}
\end{center}
\caption{The comparison between DETR and our method on ScanNet V2. \textit{KPS} represent \textit{k-Closest Points Sampling}, \textit{iter pred} represents iterative prediction.}
\label{tab::comparison_with_detr}
\vspace{-1.5em}
\end{table}

\subsection{Comparison with DETR}
\label{sec:DETR}
DETR~\cite{carion2020detr} is a pioneer work that applies the Transformer to 2D object detection. Compared with DETR, our method involves more domain knowledge, such as the data-dependent initial object candidate generation, where DETR uses a data-independent object prior to representing each object candidate and is automatically learned without explicit supervision. Moreover, there is no iterative refinement on spatial encodings in DETR as in our approach. We evaluate these differences in 3D object detection. For a fair comparison, the backbone and decoder heads used in DETR are the same as in ours. We carefully tune the hyper-parameters for DETR and chose the best setting in comparison.

The results are shown in Table~\ref{tab::comparison_with_detr}. With the same training length of 400 epochs, DETR achieves 39.6 mAP@0.25 and 21.4 mAP@0.5, significantly worse than our method. We guess it is mainly because of optimization difficulty by the data-independent object representation. The fixed spatial encoding also may contribute to inferior performance. In fact, the performance can be improved significantly by bridging these differences, reaching 59.9 mAP@0.25 and 42.9 mAP@0.5 using the same training epochs, and 61.8 mAP@0.25 and 45.2 mAP@0.5 by longer training. 

The remaining performance gap is due to the difference in ground-truth assignments, where DETR adopts a set loss to automatically determine the assignments by detection losses and our approach manually assigns object candidates to ground-truths. This assignment may also be difficult for a network to learn.

\begin{figure}
    \centering
    \vspace{-1em}
    \includegraphics[width=1.0\columnwidth]{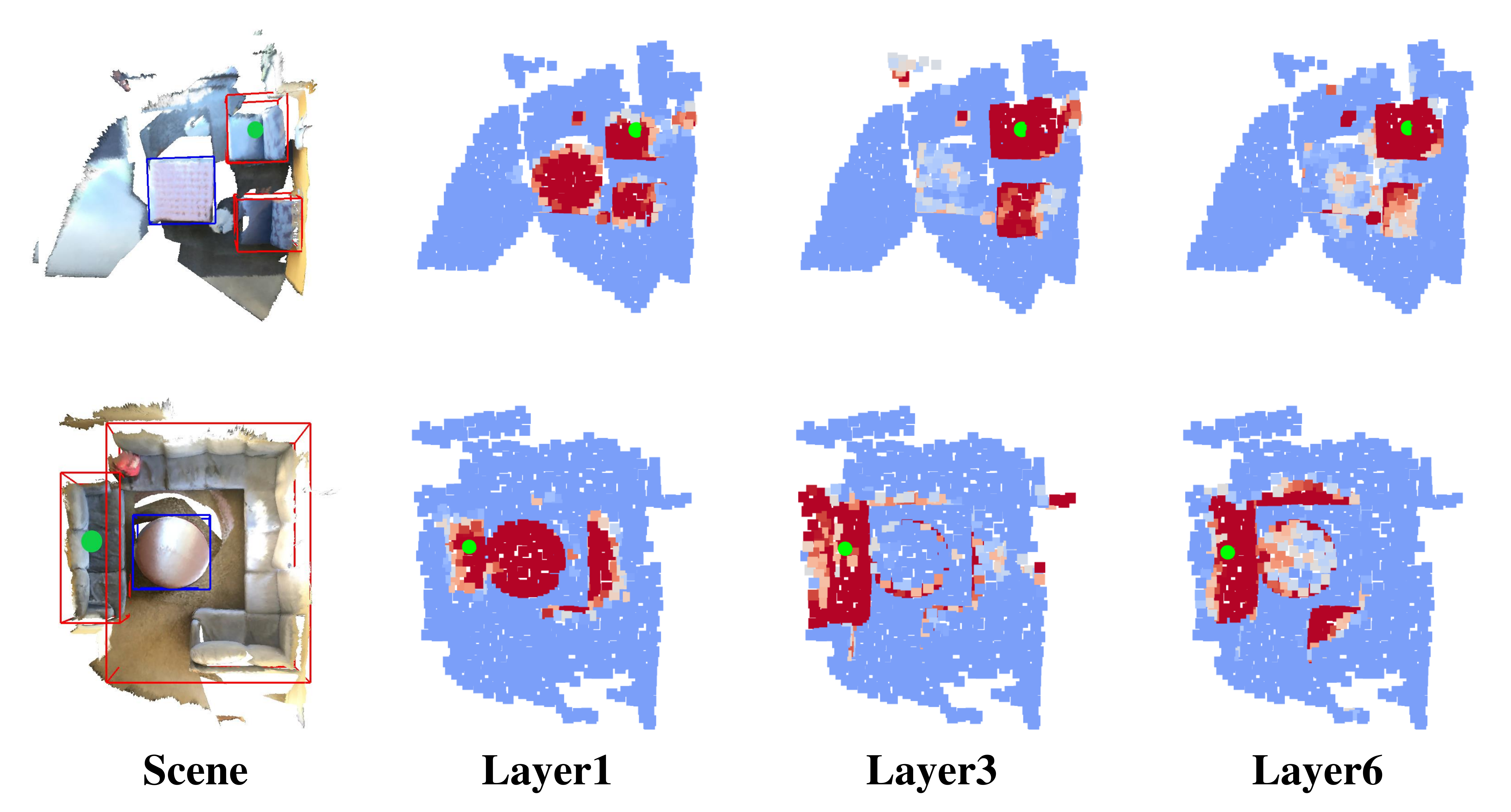}
    \caption{Visualizations on cross-attention weight in different decoder stages. The green point represents the reference object candidates. The redder color represent higher attention weight.}
    \label{fig:attention_weight}
    \vspace{-1em}
\end{figure}

\subsection{Qualitative Results}
Fig.~\ref{fig:qualitative} illustrates the qualitative results on both ScanNet V2 and SUN RGB-D. As the decoder networks go deeper, the more accurate detection results are observed.

Fig.~\ref{fig:attention_weight} visualizes the learned cross-attention weights of different decoder stages. We could observe that the model of the lower stage always focuses on the surrounding points without considering the geometry. With the refinement, the model of the higher stage could focus more on the geometry and extract more high-quality object features.

\section{Conclusion}
In this paper, we present a simple yet effective 3D object detector based on the attention mechanism in Transformers. Unlike previous methods that require a grouping step for object feature computation, this detector is group-free which computes object features from all points in a point cloud, with the contribution of each point automatically determined by the attention modules. The proposed method achieves state-of-the-art performance on ScanNet V2 and SUN RGB-D benchmarks.

\appendix

\renewcommand{\thesection}{A\arabic{section}}

\section{Training Details}

\subsection{Our Approach}
\paragraph{ScanNet V2}
We follow recent practice~\cite{qi2019votenet,xie2020mlcvnet} to use the PointNet++ as our default backbone network for a fair comparison. The backbone network has four set abstraction layers and two feature propagation layers. For each set abstraction layer, the input point cloud is sub-sampled to 2048, 1024, 512, and 256 points with the increasing receptive radius of 0.2, 0.4, 0.8, and 1.2, respectively. Then, two feature propagation layers successively up-sample the points to 512 and 1024, respectively.

In the training phase, we use 50k\footnote{We evaluate our model on 40k points on ScanNet V2 according to previous works and the performance is similar: 66.3(40k) vs. 66.2(50k) on mAP@0.25, and 48.5(40k) vs. 48.6(50k) on mAP@0.5.} points as input and adopt the same data augmentation as in \cite{qi2019votenet}, including a random flip, a random rotation between [$-5^{\circ}$, $5^{\circ}$], and a random scaling of the point cloud by [0.9, 1.1]. The network is trained from scratch by the AdamW optimizer ($\beta_1$=0.9, $\beta_2$=0.999) with 400 epochs. The weight decay is set to 5e-4. The initial learning rate is 0.006 and decayed by 10$\times$ at the 280-th epoch and the 340-th epoch. The learning rate of the attention modules is set as 1/10 of that in the backbone network. The \textit{gradnorm\_clip} is applied to stabilize the training dynamics. Following~\cite{qi2019votenet} we use class-aware head for box size prediction.

\paragraph{SUN RGB-D}
The implementation settings mostly follow~\cite{qi2019votenet}. We use 20k points as input for each point cloud. The network architecture and the data augmentation are the same as that for ScanNet V2. As the orientation of the 3D box is required in evaluation, we include an additional orientation prediction branch for all decoder layers. The orientation branch contains a classification task and an offset regression task with loss weights of 0.1 and 0.04, respectively. 

In training, the network is trained from scratch by the AdamW optimizer ($\beta_1$=0.9, $\beta_2$=0.999) with 600 epochs if not specified. The initial learning rate is 0.004 and decayed by 10$\times$ at the 420-th epoch, the 480-th epoch, and the 540-th epoch. The learning rate of attention modules is set as 1/20 of the backbone network. The weight decay is set to 1e-7, and the \textit{gradnorm\_clip} is used. We use class-agnostic head for size prediction.

\subsection{Other Pooling Mechanisms}
For a fair comparison, we only switch the feature aggregation mechanism while all other settings remain unchanged. In the following, we will introduce the implementation details of RoI-Pooling and Voting aggregation mechanism.

\paragraph{RoI-Pooling}
For a given object candidate, the points within the predicted box of the object candidate are aggregated together, and the refined box is predicted from the aggregated features. The same as our group-free approach, the multi-stage refinement is also adopted. Thus the aggregated points and features will be updated and refined in multiple stages.
Also, we tried two different strategies for feature aggregation: average-pooling and max-pooling. The results are shown in Table.~\ref{tab::ablation_roi_pooling}. We could find that the approach with max-pooling performs better, so we use it for comparison by default. 

\begin{table}
\begin{center}
\begin{tabular}{c|c|c}
\hline
method & mAP@0.25 & mAP@0.5 \\
\hline
\hline
average & 64.2 & 44.2 \\
max & 65.1 & 44.4 \\
\hline
\end{tabular}
\end{center}
\caption{Comparison between average-pooling and max-pooling on ScanNet V2.}
\label{tab::ablation_roi_pooling}
\vspace{-1em}
\end{table}

\paragraph{Voting}
The voting mechanism is first introduced by VoteNet~\cite{qi2019votenet} and we implement it in our framework. Specifically, each point predicts the center of its corresponding object, and if the distance between the predicted center of points and the center of an object candidate is less than a threshold (set to 0.3 meters), then these points and the candidate are grouped. Further, a two-layer MLP with max-pooling is used to form the aggregation feature of the object candidate, and the refined boxes are predicted from the aggregated features in the multi-stage refinement process.

\begin{table*}[htb!]
\begin{center}
\footnotesize
\setlength{\tabcolsep}{2pt}
\begin{tabular}{l|c|cccccccccccccccccc|c}
\toprule
methods & backbone & cab & bed & chair & sofa & tabl & door & wind & bkshf & pic & cntr & desk & curt & fridg & showr & toil & sink & bath & ofurn & mAP \\ 
\midrule
VoteNet~\cite{qi2019votenet} & PointNet++ & 47.7& 88.7 & 89.5 & 89.3 & 62.1 & 54.1 & 40.8 & 54.3 & 12.0 & 63.9 & 69.4 & 52.0 & 52.5 & 73.3 & 95.9 & 52.0 & 92.5 & 42.4 & 62.9 \\
MLCVNet~\cite{xie2020mlcvnet} & PointNet++ & 42.5 & 88.5 & 90.0 & 87.4 & 63.5 & 56.9 & 47.0 & 56.9 & 11.9 & 63.9 & 76.1 & 56.7 & \textbf{60.9} & 65.9 & 98.3 & 59.2 & 87.2 &  47.9 & 64.5 \\
H3DNet~\cite{zhang2020h3dnet}& 4$\times$PointNet++ & 49.4 &88.6 &91.8 &\textbf{90.2} &64.9 &\textbf{61.0} &51.9& 54.9 &\textbf{18.6} &62.0 &75.9 &57.3 &57.2 &75.3 &97.9 &67.4 &92.5 &53.6 &67.2 \\
\midrule
Ours (L6, O256) & PointNet++ & 54.1 & 86.2& 92.0 & 84.8 & 67.8 & 55.8 & 46.9 & 48.5 & 15.0 & 59.4 & 80.4 & 64.2 & 57.2 &\textbf{76.3} & 97.6 & \textbf{76.8} & 92.5 & 55.0 & 67.3 \\
Ours (L12, O256) & PointNet++ & 55.4 & 86.6 & 91.8 & 86.6 & \textbf{73.0} & 54.5 & 49.4 & 47.7 & 13.1 & 63.3 & \textbf{82.4}& 63.3& 53.2 & 74.0 & 99.2 & 67.7 & 91.7 & 55.8 &67.2  \\
Ours (L12, O256) & PointNet++w2$\times$ & \textbf{56.5} &88.2 & 92.5 & 88.2 & 71.6 & 57.5 & 48.3 & 53.7 & 17.5 & \textbf{71.0} & 79.5 & 63.4 & 58.1 & 71.7 & 99.4 & 71.1 & 93.0 & \textbf{57.8} & 68.8 \\
Ours (L12, O512) & PointNet++w2$\times$ & 52.1 & \textbf{91.9} & \textbf{93.6} & 88.0 & 70.7 & 60.7 & \textbf{53.7} & \textbf{62.4} & 16.1 & 58.5 & 80.9 & \textbf{67.9} & 47.0 & \textbf{76.3} & \textbf{99.6} & 72.0 & \textbf{95.3} &56.4  & \textbf{69.1} \\

\bottomrule
\end{tabular}
\end{center}
\caption{Performance of mAP@0.25 for each category on the ScanNet V2 dataset.}
\label{tab:perclassscannet025}
\end{table*}

\begin{table*}[htb!]
\begin{center}
\footnotesize
\setlength{\tabcolsep}{2pt}
\begin{tabular}{l|c|cccccccccccccccccc|c}
\toprule
methods & backbone & cab & bed & chair & sofa & tabl & door & wind & bkshf & pic & cntr & desk & curt & fridg & showr & toil & sink & bath & ofurn & mAP \\ 
\midrule
VoteNet~\cite{qi2019votenet} & PointNet++ & 14.6 & 77.8 & 73.1 & \textbf{80.5} & 46.5 & 25.1 & 16.0 & 41.8 & 2.5 & 22.3 & 33.3 & 25.0 & 31.0 & 17.6 & 87.8 & 23.0 & 81.6 & 18.7 & 39.9 \\
H3DNet~\cite{zhang2020h3dnet}& 4$\times$PointNet++ & 20.5 &79.7& 80.1& 79.6& 56.2& 29.0& 21.3& 45.5 & 4.2& 33.5& 50.6& 37.3& 41.4& 37.0& 89.1& 35.1& \textbf{90.2}& 35.4 & 48.1 \\
\midrule
Ours (L6, O256) & PointNet++ & 23.0 &78.4 &78.9 &68.7 &55.1 & 35.3 & 23.6 & 39.4 & 7.5 & 27.2 & 66.4 & 43.3& 43.0 & 41.2 & 89.7 & 38.0 & 83.4 & 37.3 & 48.9 \\
Ours (L12, O256) & PointNet++ & 23.8 & 77.2 & 81.6 & 65.1 & \textbf{62.8} & 35.0 & 21.3 & 39.4 & 7.0 & 33.1 & 66.3 & 39.3 & 43.9 & \textbf{47.0} & 91.2 & 38.5 & 85.2 & 37.4 & 49.7 \\
Ours (L12, O256) & PointNet++w2$\times$ & \textbf{26.2} & 80.7 & \textbf{83.5} & 70.7 & 57.0 & 37.4 & 21.2 & 47.7 & \textbf{8.8} & \textbf{45.3} & 60.7 & 42.2 & 43.5 & 42.7 & \textbf{95.5} & \textbf{42.3} & 89.7 & \textbf{43.4} & 52.1 \\
Ours (L12, O512) & PointNet++w2$\times$ & 26.0 & \textbf{81.3} & 82.9 & 70.7 & 62.2 & \textbf{41.7} & \textbf{26.5} & \textbf{55.8} & 7.8 & 34.7 & \textbf{67.2} & \textbf{43.9}& \textbf{44.3} &44.1 & 92.8 & 37.4 & 89.7 & 40.6 & \textbf{52.8} \\

\bottomrule
\end{tabular}
\end{center}
\caption{Performance of mAP@0.5 for each category on the ScanNet V2 dataset.}
\label{tab:perclassscannet05}
\end{table*}

\begin{table*}[htb!]
\small
\setlength{\tabcolsep}{3pt}
\begin{center}
\begin{tabular}{l|c|cccccccccc|c}
\toprule
methods & backbone & bathtub & bed & bkshf & chair & desk & drser & nigtstd & sofa & table & toilet & mAP \\
\midrule
VoteNet~\cite{qi2019votenet} & PointNet++& 75.5 & 85.6 & 31.9 & 77.4 & 24.8 & 27.9 & 58.6 & 67.4 & 51.1 & 90.5 & 59.1\\
MLCVNet~\cite{xie2020mlcvnet} & PointNet++ &  79.2 & 85.8 & 31.9 & 75.8 & 26.5 & 31.3 & 61.5 & 66.3 & 50.4 & 89.1 & 59.8 \\ 
HGNet~\cite{chen2020hierarchical} & PointNet++ w/ FPN & 78.0 &84.5&  \textbf{35.7}& 75.2&  \textbf{34.3} & 37.6& 61.7& 65.7& 51.6& \textbf{91.1}& 61.6 \\
H3DNet~\cite{zhang2020h3dnet}& 4$\times$PointNet++ & 73.8 &85.6 &31.0& 76.7 &29.6& 33.4& 65.5& 66.5& 50.8& 88.2& 60.1 \\
\midrule
Ours (L6, O256) & PointNet++ & \textbf{80.0} & \textbf{87.8} & 32.5  & \textbf{79.4} &32.6 & \textbf{36.0} & \textbf{66.7}  & \textbf{70.0}  & \textbf{53.8}  &\textbf{91.1}  & \textbf{63.0}\\ 

\bottomrule
\end{tabular}
\end{center}
\caption{Performance of mAP@0.25 for each category on the SUN RGB-D validation set.}
\label{tab:perclasssunrgbd025}
\end{table*}

\begin{table*}[htb!]
\small
\setlength{\tabcolsep}{3pt}
\begin{center}
\begin{tabular}{l|c|cccccccccc|c}
\toprule
methods & backbone & bathtub & bed & bkshf & chair & desk & drser & nigtstd & sofa & table & toilet & mAP \\
\midrule
VoteNet~\cite{qi2019votenet} & PointNet++ & 45.4 & 53.4 & 6.8 & 56.5 & 5.9 & 12.0 & 38.6 & 49.1 & 21.3 & 68.5 & 35.8 \\
H3DNet~\cite{zhang2020h3dnet}& 4$\times$PointNet++ & 47.6 &52.9 &8.6& 60.1 & 8.4& 20.6& 45.6& 50.4& 27.1& 69.1& 39.0 \\
\midrule
Ours (L6, O256) & PointNet++ & \textbf{64.0} & \textbf{67.1} & \textbf{12.4} & \textbf{62.6} &\textbf{14.5} & \textbf{21.9}& \textbf{49.8}& \textbf{58.2} & \textbf{29.2} & \textbf{72.2} & \textbf{45.2}\\

\bottomrule
\end{tabular}
\end{center}
\caption{Performance of mAP@0.5 for each category on the SUN RGB-D validation set.} 
\label{tab:perclasssunrgbd05}
\end{table*}

\section{More Results}
We show per-category results on ScanNet V2 and SUN RGB-D under different IoU thresholds. Table~\ref{tab:perclassscannet025} and Table~\ref{tab:perclassscannet05} show the results of mAP@0.25 and mAP@0.5 on ScanNet V2, respectively. Table~\ref{tab:perclasssunrgbd025} and Table~\ref{tab:perclasssunrgbd05} show the results of mAP@0.25 and mAP@0.5 on SUN RGB-D, respectively.

\begin{figure*}[htb!]
    \centering
    \includegraphics[width=0.6\linewidth]{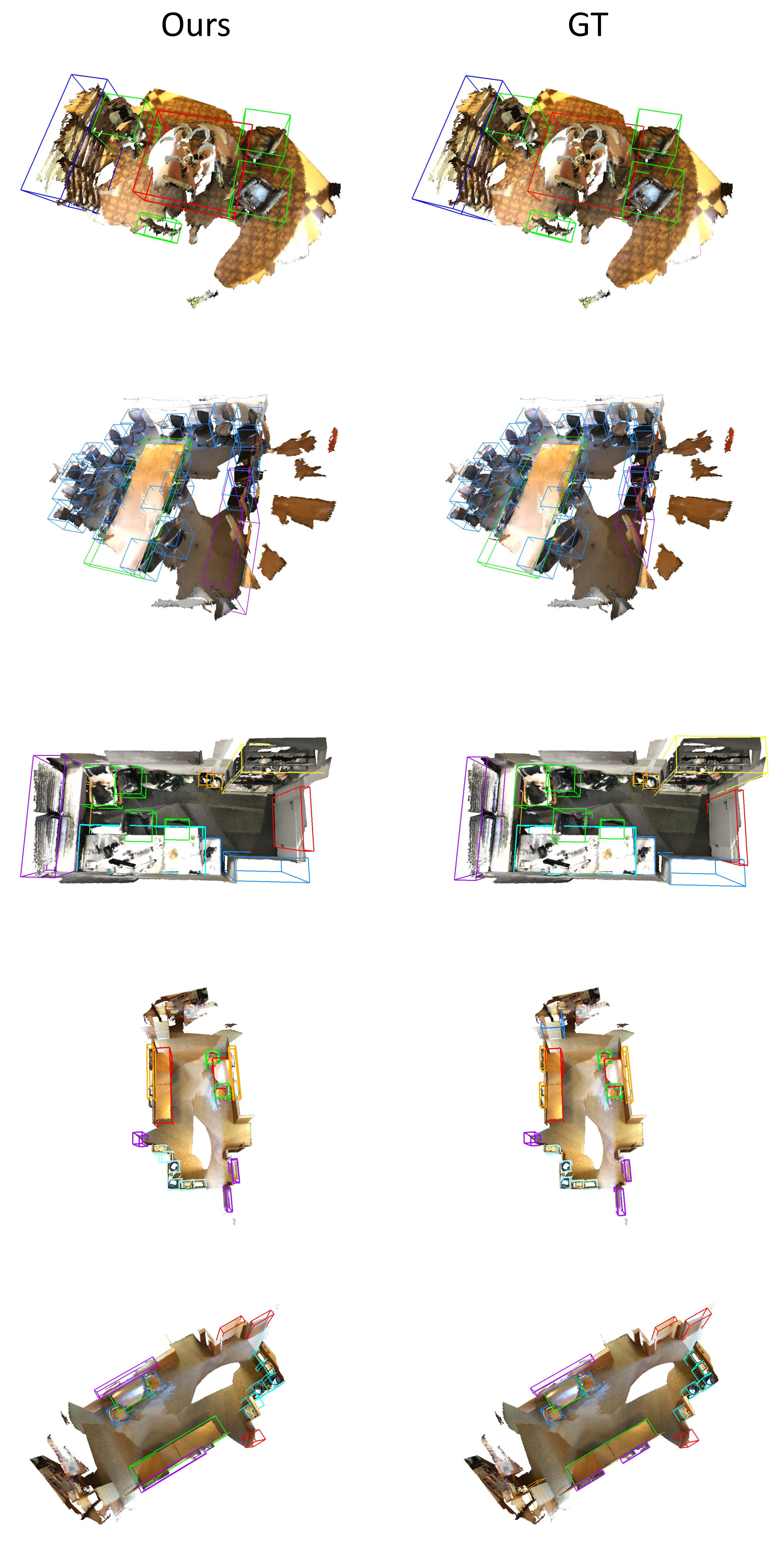}
    \caption{Qualitative results on ScanNet V2.}
    \label{fig:scannet_qualitative_more}
\end{figure*}

\begin{figure*}[htb!]
    \centering
    \includegraphics[width=0.8\linewidth]{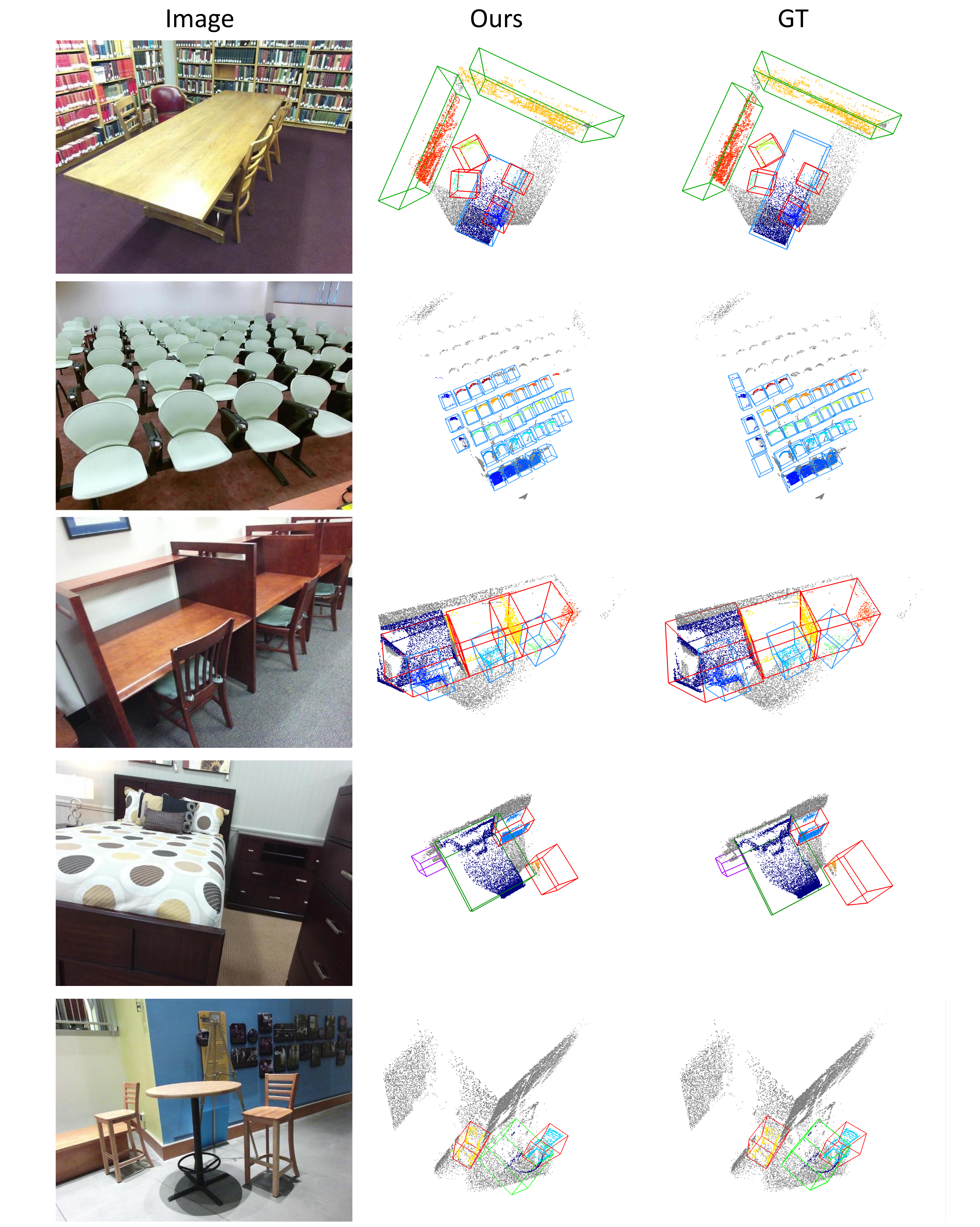}
    \caption{Qualitative results on SUN RGB-D.}
    \label{fig:sunrgbd_qualitative_more}
\end{figure*}

We also show more qualitative results of our method on ScanNet V2 and SUN RGB-D. The results are shown in Figure~\ref{fig:scannet_qualitative_more} (ScanNet V2) and Figure~\ref{fig:sunrgbd_qualitative_more} (SUN RGB-D).

{\small
\bibliographystyle{ieee_fullname}
\bibliography{egbib}
}

\end{document}